\documentclass[11pt,a4paper]{article}

\usepackage{authblk}
\usepackage[hyperref]{acl2020}
\usepackage{times}
\usepackage{latexsym}
\usepackage{graphicx}
\usepackage{multirow}
\usepackage{tabularx}
\usepackage{makecell}

\newcolumntype{H}{>{\setbox0=\hbox\bgroup}c<{\egroup}@{}}

\newcommand{\specialcell}[2][l]{%
  \begin{tabular}[#1]{@{}c@{}}#2\end{tabular}}

\usepackage{microtype}

\newenvironment{itemize2}
    {\begin{itemize}
        \vspace{-0.3em}
        \setlength{\abovedisplayskip}{0pt}
        \setlength{\belowdisplayskip}{0pt}
        \setlength{\itemsep}{5pt}
        \setlength{\parskip}{0pt}
        \setlength{\parsep}{0pt}
        \setlength{\topsep}{0pt}
        \setlength{\partopsep}{0pt}
    }
    {\vspace{-0.3em}
    \end{itemize}}

\newenvironment{enumerate2}
    {\begin{enumerate}
        \vspace{-0.1em}
        \setlength{\abovedisplayskip}{0pt}
        \setlength{\belowdisplayskip}{0pt}
        \setlength{\itemsep}{5pt}
        \setlength{\parskip}{0pt}
        \setlength{\parsep}{0pt}
        \setlength{\topsep}{0pt}
        \setlength{\partopsep}{0pt}
    }
    {\vspace{-0.1em}
    \end{enumerate}}

\aclfinalcopy 

\title{A Comparative Study of Lexical Substitution Approaches\\based on Neural Language Models}

\author[$\dag,\ddag$]{\textbf{Nikolay Arefyev}}
\author[$\dag$]{\textbf{Boris Sheludko}}
\author[$\dag$]{\textbf{Alexander Podolskiy}}
\author[$\star$]{\textbf{Alexander Panchenko}}

\affil[$\dag$]{Samsung R\&D Institute Russia, Moscow, Russia}

\affil[$\ddag$]{National Research University Higher School of Economics, Moscow, Russia}

\affil[$\star$]{Skolkovo Institute of Science and Technology, Moscow, Russia}

\date{}

\begin{document}
\maketitle
\begin{abstract}
Lexical substitution in context is an extremely powerful technology that can be used as a backbone of various NLP applications, such as word sense induction, lexical relation extraction, data augmentation, etc. In this paper we present a large-scale comparative study of popular neural language and masked language models (LMs and MLMs), such as context2vec, ELMo, BERT, XLNet, applied to the task of lexical substitution. We show that already competitive results achieved by SOTA LMs/MLMs can be further improved if information about the target word is injected properly, and compare several target injection methods. In addition, we provide analysis of the types of semantic relations between the target and substitutes generated by different models providing insights into what kind of words are really generated or given by annotators as substitutes. 

\end{abstract}

\section{Introduction}

Lexical substitution is the task of generating words which can replace a given word in a given textual context. For instance, in the sentence ``My daughter purchased a new \textit{car}'' the word \textit{car} can be substituted by its synonym \textit{vehicle} keeping the same meaning, but also with the co-hyponym \textit{bike}, or even the hypernym \textit{means of transport} while keeping the original sentence grammatical. Lexical substitution can be useful in various applications, such as word sense induction~\cite{amrami-2018}, lexical relation extraction~\cite{schick2019rare}, paraphrase generation, semantic spelling correction, text simplification, textual data augmentation, etc.

The new generation of language models (LMs) based on deep neural networks, such as  ELMo~\cite{peters-etal-2018-deep}, BERT~\cite{devlin2018pretraining}, and XLNet~\cite{yang2019xlnet} enabled a profound breakthrough in many NLP tasks, ranging from sentiment analysis to named entity recognition. Commonly these models are used to perform pre-training of deep neural networks which are finally fine-tuned to perform some task different from language modelling~\cite{howard-ruder-2018-universal}. In this paper we provide the first large-scale comparison and analysis of various neural LMs/MLMs applied to the task of lexical substitution and two tasks which exploit lexical substitution, namely word sense induction and text data augmentation. More specifically, the main contributions of the paper are as follows:

\begin{itemize2}
\item A \textit{comparative study} of neural language models and masked language models (context2vec, Elmo, BERT, XLNet) applied for lexical substitution based on both intrinsic and extrinsic evaluations. 
\item A \textit{study of types of semantic relations} (synonyms, co-hyponyms, etc.) produced by substitution models and human annotators. 
\item A \textit{study of methods of target word inclusion} for improvement of lexical substitution quality. 
\end{itemize2}

\section{Related Work}
\label{sec:rel_work}

The paper which is arguably most similar to our study is~\cite{zhou-etal-2019-bert}, where an end-to-end lexical substitution approach based on BERT is proposed, similar to the baseline BERT-based approaches studied in our paper. However, our study goes beyond evaluation only on the SemEval-based lexical substitution task: in addition to this, we test performance on other intrinsic datasets but also in the context of two applications: word sense induction and data augmentation. Besides, our study is not limited to BERT but compares face-to-face three recently introduced neural LMs: BERT, ELMo, and XLNet and their variants. 

More generally, solving the lexical substitution task requires finding words that are both appropriate in the given context and related to the target word in some sense (which may vary depending on the application of generated substitutes). To achieve this unsupervised substitution models heavily rely on distributional similarity models of words (DSMs) and language models (LMs). Probably, the most commonly used DSM is \textit{word2vec} model, which trains word embeddings and context embeddings to be similar when they tend to occur together. Contexts are either nearby words \cite{word2vec}, or syntactically related words \cite{levy-goldberg-2014-dependency}, resulting in similar embeddings for distributionally similar words. In \cite{melamud-etal-2015-simple} several metrics for lexical substitution were proposed based on embedding similarity of substitutes both to the target word and to the words in the given context. Later \cite{pic} improved this approach by switching to dot-product instead of cosine similarity and applying an additional trainable transformation to context word embeddings. 

A more sophisticated \textit{context2vec} model producing embeddings for a word in a particular context (contextualized word embeddings) was proposed in \cite{c2v} and was shown to outperform previous models in a ranking scenario when candidate substitutes are given. The training objective is similar to \textit{word2vec}, but context representation is produced by two LSTMs (a forward and a backward for the left and the right context), in which final outputs are combined by a feed-forward NN. For lexical substitution, candidate word embeddings are ranked by their similarity to the given context representation. A similar architecture consisting of a forward and a backward LSTM is employed in \textit{ELMo} \cite{peters-etal-2018-deep}. However, each LSTM was trained with the LM objective instead. To rank candidate substitutes using \textit{ELMo} \cite{soler-etal-2019-comparison} proposed calculating cosine similarity between contextualized \textit{ELMo} embeddings of the target word and all candidate substitutes (this requires feeding the original example with the target word replaced by one of the candidate substitutes at a time). Average of all \textit{ELMo} layers' outputs at the target timestep performed best. However, they found \textit{context2vec} perform even better explaining this by its training objective, which is more related to the task.

Recently deep Transformer NNs pre-trained on huge corpora with LM or similar objective consistently show SOTA results in a variety of NLP tasks. BERT \cite{devlin2018pretraining} is trained to restore a word replaced with a special [MASK] token at its input given both left and right context (masked LM objective) and \textit{XLNet} \cite{yang2019xlnet} predicts a word at a specified position given randomly selected words from the context with their positions. In \cite{zhou-etal-2019-bert}, \textit{BERT} was reported to perform purely for lexical substitution (which is contrary to our experiments) and two improvements were proposed to achieve SOTA results using it. Firstly, dropout is applied to the target word embedding before showing it to the model. Secondly, the similarity between the original contextualized representations of context words and their representations after replacing the target by one of the possible substitutes are integrated into the ranking metric to ensure minimal changes in the sentence's meaning. We are not aware of any work applying \textit{XLNet} for lexical substitution, but our experiments show that it outperforms \textit{BERT} by a large margin.

Supervised approaches to lexical substitution were also proposed, including \cite{szarvas-etal-2013-supervised,szarvas-etal-2013-learning,hintz-biemann-2016-language}. These approaches rely on manually curated lexical resources like WordNet, so they are not easily transferrable to different languages unlike those described above. Also, the latest unsupervised methods like \cite{zhou-etal-2019-bert} were shown to perform better.

\section{Neural Language Models for Lexical Substitution}
\label{sec:methods}

To generate a substitute we take a text fragment and a target word position in it as input, and produce a list of substitutes with their probabilities using a neural LM/MLM. We experiment with naive application of MLMs to predict probability distribution for words that can appear instead of the target word given its left and right context, and also with combinations of several probability distributions including distributional similarity to the target. Combinations yield better results for WSI according to prior studies \cite{amrami-2019,ranlp2019} and further for intrinsic and extrinsic metrics in our experiments. More specifically, various methods for inclusion information about the target word are tested. 

In our experiments, we the following models as substitute probability estimators: context2vec~\cite{c2v}, ELMo~\cite{peters-etal-2018-deep}, BERT~\cite{devlin2018pretraining}, and XLNet~\cite{yang2019xlnet}.
Our experiments show that neural LMs used for lexical substitutes should preserve the meaning of the target word: the information about the target should be somehow presented to the substitute generator. This is why we experiment with several ways to inject information about the target word. Suppose we have an example $LTR$, where $T$ is the target word, and $C = (L, R)$ is its context (left and right correspondingly).

The first option is to combine a distribution provided by substitute probability estimator, $P(s|C)$, with a distribution that comes from measuring of proximity between the target and substitutes, $P(s|T)$. The latter distribution is computed as an inner product between the respective embeddings. If we simply multiply these distributions the second will almost have no effect because the first is very peaky. To align the orders of distributions we use softmax with temperature:
 $   P(s|T) \propto \exp(\frac{\langle emb_s, emb_T\rangle}{\mathcal{T}}).$
The final distribution is obtained by the formula
   $ P(s|C, T) \propto \frac{P(s|C)P(s|T)}{P(s)^\beta},$
where $\beta$ is a parameter controlling how we penalize frequent words, for more details see \cite{ranlp2019}.

The second option is to use dynamic patterns. For example, pattern ``T and then $\_$'', proposed in \cite{arefyev-semeval-2019} means that we replace the target with this construction. The probability estimator should predict words at timestep ‘\_’. Dynamic patterns give a vision of the target word to the model.

Finally, we can give no information about the target word to the probability estimator. By default, ELMo does not have this information. BERT has special {\it mask} tokens, so we replace the target word with this token, thus, hiding the target from the model. For XLNet we use special attention mask so words in the context don't see the target word. 

More specifically, we experiment with the following baseline models and their upgraded version which include one of these approaches:

\paragraph{ELMo} To use ELMo as a probability estimator divide a sentence into left and right contexts with respect to a target word. We obtain two independent distributions over vocabulary: one with the forward model for the left context, $P(s|L)$, another with the backward model for the right context, $P(s|R)$. To combine these distributions by using method BComb-LMs proposed in \cite{ranlp2019}. Therefore, we get distribution: $P(s|L,R) = \frac{P(s|L)P(w|R)}{P^{\beta}(s)}$. The substitutes are the most probable words according to this distribution. Additionally, we study this model with two types of target injection: proximity according to ELMo-embeddings,  denoted as \textit{ELMo+embs}, and dynamic-patterns usage, denoted as \textit{ELMo+pat}.

\paragraph{BERT} In order to generate substitutes with BERT we give full context as input to a model and gather distribution over vocabulary at target word position. Since BERT is a masked LM we can mask out target word, hence, using no target word information to a model. Such a generator we would call \textit{BERT-notgt}. As for ELMo we furthermore analyze other target word injections: combination with first layer BERT embeddings (\textit{BERT+embs}) and combination with dynamic-pattern (\textit{BERT+pat}).

\paragraph{XLNet} We obtain substitute distribution with XLNet in the same way as for BERT. In the case of a base model, elements at context positions could attend to an element at a target position, non-masked version. In a similar vein as for BERT and ELMo we consider three additional models: combination with embeddings (\textit{XLNet+embs}), masking of a target word (\textit{XLNet-notgt}), and usage of dynamic-pattern  (\textit{XLNet+pat}). We find that for small contexts XLNet gives erroneous distribution. To mitigate this problem we prepend initial context with some text that ends with the end of document special symbol.

\section{Baseline Lexical Substitution Models}

Lexical substitution models based on the three state-of-the-art neural LMs described above are compared to the three following strong models specifically developed for the lexical substitution task: OOC~\cite{pic}, nPIC~\cite{pic}, and context2vec \cite{c2v}.

\paragraph{OOC: Out of Context}

This model ranks words by their cosine similarity with the target word and completely ignores context. Following \cite{pic} we use dependency-based embeddings\footnote{\url{http://www.cs.biu.ac.il/nlp/resources/downloads/lexsub_embeddings}} released by \cite{melamud-etal-2015-simple}.

\paragraph{nPIC: non-Parameterized probability In Context}

nPIC is a measure that consists of two independent components that measure appropriateness of a substitute to the context (words that are directly connected to the target) and to the target, see \cite{pic}. Each component is based on dependency based word and context embeddings and takes form of a softmax.

\paragraph{context2vec}

This model builds the vector representation of the context using LSTM-based NN and ranks possible substitutes by their dot product similarity to the context representation. We use original implementation\footnote{\url{https://github.com/orenmel/context2vec}} and the weights\footnote{\url{http://u.cs.biu.ac.il/~nlp/resources/downloads/context2vec}} pre-trained on ukWac dataset.

\begin{table*}[!t]
\footnotesize 
\begin{center}
\begin{tabular}{|l|c|c|c|c|c|c|c|c|}
\hline
\multirow{2}{*}{Model} & \multicolumn{4}{c|}{SemEval 2007} & \multicolumn{4}{c|}{CoInCo} \\ \cline{2-9} 
            & GAP  & P@1  & P@3  & R@10 & GAP  & P@1  & P@3  & R@10 \\ \hline
OOC         & 42.8 & 15.9 & 12.5  & 18.1    & 44.5 & 10.9 & 8.6  & 14.3  \\ \hline
nPIC        & 50.6 & 23.1 & 17.3 & 26.5 & 48.1 & 26.3 & 19.8 & 18.0  \\ \hline

context2vec & 53.4 & 7.6 & 5.6 & 10.8 & 47.5 & 8.4  & 7.0 & 7.9  \\ \hline
\hline

ELMo & 51.7 & 11.4 & 8.4  & 14.0 & 48.9 & 13.6 & 11.0 & 11.6  \\ \hline
BERT & 52.8 & 37.1 & 27.0 & 38.8 & 50.3 & 44.4 & 33.6 & 29.8  \\ \hline
XLNet & 57.0 & 31.3 & 22.4 & 34.2 & 52.8 & 39.2 & 29.4 & 27.2  \\ \hline
ELMo+embs & 53.6 & 33.8 & 23.3 & 33.8 & 53.3 & 38.4 & 28.7 & 26 \\ \hline
BERT+embs & 52.1 & 38.5 & 29.4 & 42.6 & 50.3 & 44.8 & 35.2 & 31.9   \\ \hline
XLNet+embs &  \textbf{57.3} & \textbf{45.6} & \textbf{32.8} & \textbf{46.0} & \textbf{54.8} & \textbf{49.0} & \textbf{38.9} & \textbf{34.9}  \\ \hline

\end{tabular}
\caption{Results for candidate ranking (GAP) and all words ranking based on our re-implementation of the baselines to ensure that the models use the same post-processing.}
\label{tab:eval_res}
\end{center}
\end{table*}

\section{Intrinsic Evaluation}
\label{sec:intrinsic_eval}

We perform an intrinsic evaluation of neural LMs  on the lexical substitution task on two datasets. 

\subsection{Experimental Settings}
Lexical substitution task is concerned with finding appropriate substitutes for a target word in a given context. For example, possible substitutes of a word {\bf trade} in the sentence "{\it Angels make a trade to get outfield depth.}" are a swap, exchange, deal, barter, transaction, etc. The irrelevant ones are skill or craft that encompass different meanings of {\bf trade}. This task was originally introduced as SemEval 2007 evaluation competition \cite{mccarthy-navigli-2007-semeval} and suits for an evaluation of how distributional models handle polysemous words. In a lexical substitution task, annotators are provided with the target word and the context. Their task is to propose possible substitutes. Since there are several annotators, we have a weighted list of substitutes for each target word in a given context. 

We compute the probability of a substitute for a target word in a context acquiring distribution over vocabulary or a candidate list. Lexical substitution task comes with two variations: candidate ranking and all-words ranking. In candidate ranking task models are provided with the list of {\it candidates}. Following previous works, we acquire this list by merging all the substitutions of the target lemma and POS tag over the corpus. We measure performance on this task with Generalized Average Precision (GAP) that was introduced in \cite{gap}. GAP is similar to Mean Average Precision and the difference is in the weights that come from how many times annotators selected a particular substitute (see the original paper on GAP for more details). Following \cite{melamud-etal-2015-modeling} we discard all multi-word expressions from the gold substitutes and omit all instances that left without gold substitutes. 

In all-ranking task model is not given with the candidate substitutions, therefore, it's a much harder task than the previous one. The model should give a higher probability to gold substitutes than to other words in its vocabulary that could have the size of thousands of words. Commonly data sets don't have many annotators and many words have a lot of possible substitutes, e.g. you could change word {\it violet} on many other colors. Hence, it's challenging for the model to generate substitutes that were chosen by annotators. Following \cite{pic} we use mean precision at 1 and 3 (P@1, P@3) as an evaluation metric for this task. Additionally, we look at recall at 10 (R@10).

We used two lexical substitution datasets:

\textbf{SemEval 2007 task} \cite{mccarthy-navigli-2007-semeval}  consists of 300 dev and 1710 test sentences for 201 polysemous words. For each target word, 10 sentences are provided. Annotators' task was to give up to 3 possible substitutes.
    
\textbf{CoInCo} or Concepts-In-Context dataset~\cite{kremer-etal-2014-substitutes} consists of over than 15K target instances with a given 35\%/65\% split. There are over 2500 sentences that come from fiction, emails, and newswires. Annotators provided at least 6 substitutes for each target.
    

\subsection{Discussion of Results}

\paragraph{Comparison to previously published baselines}

Table~\ref{tab:prevbest_res} contains metrics for candidate and all-vocab ranking tasks. We compare our best model (XLNet+embs) with a baseline models presented in \cite{pic}, context2vec (c2v) model \cite{c2v} and BERT for lexical substitution presented in \cite{zhou-etal-2019-bert}. Note the improvement of the proposed model over baselines. On the SemEval07 task, our models show comparable results to c2v but outperform it on the CoInCo data set. BERT for lexical substitution outperforms XLNet+embs on all tasks. In \cite{zhou-etal-2019-bert} they add substitute validation metric that improves predictions. We expect that our models also could be improved with this technique. That leaves room for future research. It is worth to mention that BERT and XLNet work on a sub-token level, hence, their vocabularies are lower in size than ELMo or c2v and contain a lot of non-word tokens. We hypothesize that these models could be improved by integrating multi-token generation so they could cover more words. 

\begin{table}[!h]
\footnotesize 
\begin{tabular}{|c|p{15pt}|c|p{15pt}|c|}
\hline
\multirow{2}{*}{Model} & \multicolumn{2}{c|}{SemEval 2007} & \multicolumn{2}{c|}{CoInCo}  \\ \cline{2-5} 
            & GAP  & P@1/@3 & GAP  & P@1/@3  \\ \hline
Sup. learning & 55.0 & - & - & -  \\ \hline
Trans. learning & 51.9 & - & - & -  \\ \hline
PIC & 52.4 & 19.7/14.8 & 48.3 & 18.2/13.8  \\ \hline
Substitute vector & 55.1 & - & 50.2 & -  \\ \hline
context2vec & 56.0 & - & 47.9 & -  \\ \hline
BERT$s_p,s_v$ & 60.5 & 51.1/- & 57.6 & 56.3/-  \\ \hline
\hline
{XLNet+embs} & 57.3 & 45.6/32.8 & 54.8 & 49.0/38.9  \\ \hline
\specialcell{XLNet+embs\\(w/o trg excl)} & 57.4 & 13.2/18.8 & 51.6 & 14.8/24.2  \\ 
\hline
\specialcell{XLNet+embs\\(w/o lemmat)} & 58.6 & 24.4/17.8 & 51.5 & 25.5/19.3  \\ \hline
\specialcell{XLNet+embs\\(c2v post-proc)} & 58.8 & 25.8/21.0 & 52.9 & 17.7/16.8  \\ \hline
\end{tabular}
\caption{Comparison to previous published results. Post-processing and metrics implementation details may differ. Models: Supervised Learning~\cite{szarvas-etal-2013-learning}, Transfer Learning~\cite{hintz-biemann-2016-language}, PIC~\cite{pic}, Substitute vector~\cite{melamud-etal-2015-modeling}, context2vec~\cite{c2v}, BERT$s_p,s_v$~\cite{zhou-etal-2019-bert}.}
\label{tab:prevbest_res}
\end{table}

Further, Table~\ref{tab:prevbest_res} provide results for different post-processing of substitute distribution from our XLNet+embs model.
We see that post-processing has a great impact on the metrics. 
In PIC authors use NLTK English stemmer for exclusion stems of the target word, i.e. they assign zero probability to word with stem equal to target stem. The code of context2vec uses NLTK WordNet lemmatizer to lemmatize only candidates. We use spaCy lemmatizer in our post-processing.
In order to analyze substitute distributions provided by different vectorizers, independently of post-processing steps, we fixed the following post-processing: default post-processing (i.e. with lemmatization and target exclusion), w/o lemmatization, w/o target lemmas exclusion, c2v post-processing. In Table~\ref{tab:eval_res} we present results for our re-implementations of baselines, context2vec and proposed generators.


\paragraph{Re-implementation of the baselines}

Table~\ref{tab:eval_res} contains metrics (P@1, P@3, R@10) for all-words ranking variation of lexical substitution task. First, we note that pipelines based on a new line of NLP models (ELMo, BERT, XLNet) substantially outperform word2vec based PIC and OOC methods. We have approximately 50\% relative improvement in precision@1 for SemEval07 and 60\% for CoInCo. This indicates that proposed models are better at capturing the meaning of a word in a context as such providing more accurate substitutes. We also note that combination with embeddings substantially improves basic models. The greatest improvement comes for XLNet model in precision and recall, e.g. for SemEval07 precision@1 improves by approximately 14\%.

\paragraph{Injection of information about target word}

\begin{figure}[!h]
\begin{center}
\includegraphics[width=0.5\textwidth]{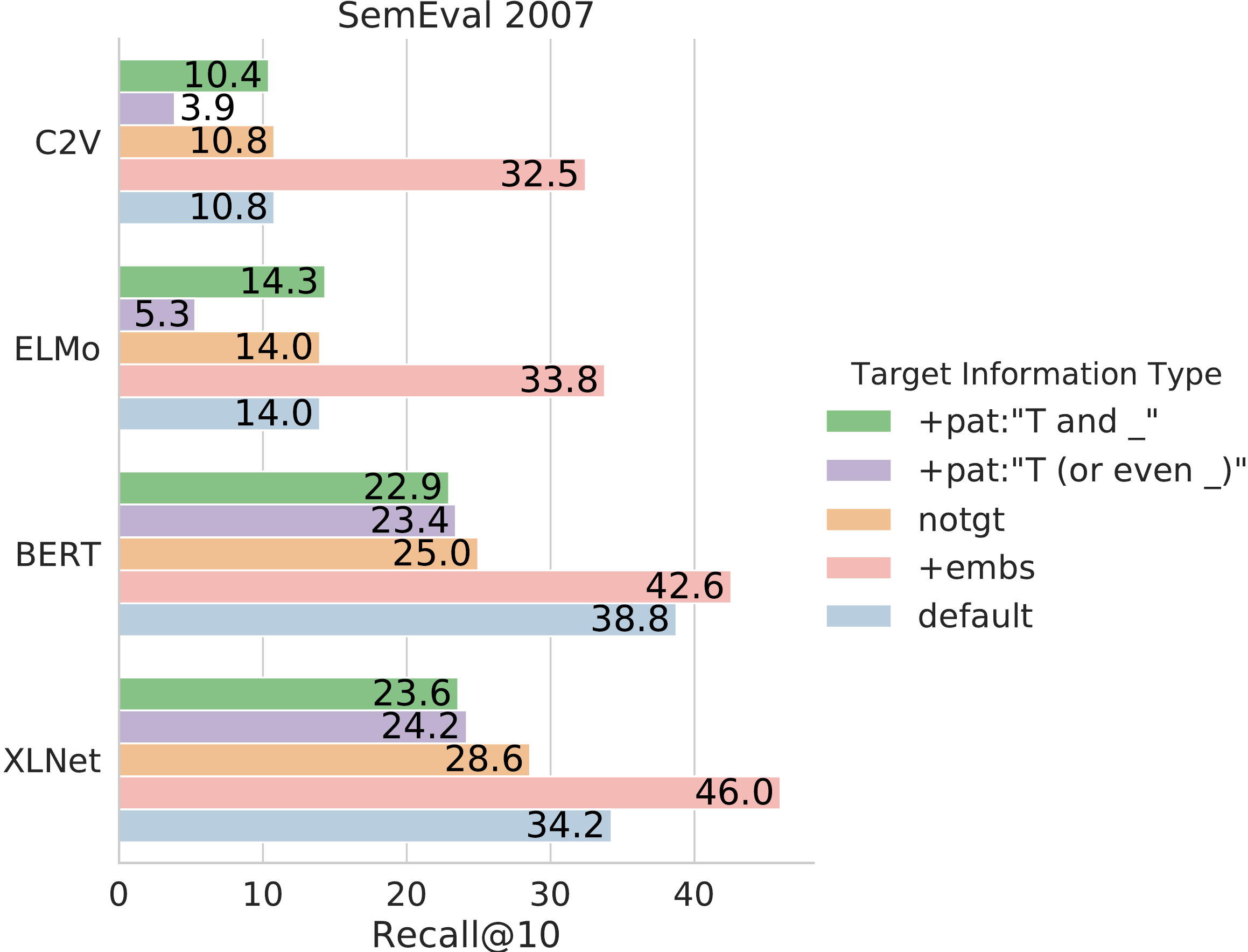}
\caption{Comparison of various target information injection methods on the SemEval 2007 dataset. By default XLNet and BERT see target and ELMo doesn't(ELMo-notgt same model as ELMo-default). }
\label{fig:target-info}
\end{center}
\end{figure}

Here we compare substitute generation models described in Section~\ref{sec:methods} using the based on different types of target information injection. Figure \ref{fig:target-info} shows Recall@10 metric on SemEval 2007 \cite{mccarthy-navigli-2007-semeval} dataset for each substitute generator. Dynamic pattern application worsens the result of XLNet-notgt and BERT-notgt generators, but ELMo with pattern 'T and \_'(proposed in \cite{amrami-2018}), slightly outperforms ELMo-notgt. Perhaps this is because people generate substitutes in the original sentence without a pattern, but in our case, despite we show target word to the substitute generator, the pattern can spoil the predictions. When we show the target word in the sentence to the substitute generator(BERT-base or XLNet-base) we overtake BERT-notgt by several percents, because the target word information allows the generator to generate more relevant substitutes. Also, the combination of a probability distribution with embedding similarity leads to a significant increase of Recall@10. For example, ELMo+embs outperforms ELMo-notgt more than 50 percent. And also XLNet+embs outperforms XLNet-base more than 12 percent. This result means that the correct information about the target word allows you to generate substitutes more similar to human substitutes and more appropriate for the context.

\begin{table}[!h]
\footnotesize
\centering
\begin{tabular}{|c|c|c|}
\hline
\multicolumn{1}{|c|}{\multirow{2}{*}{\textbf{Model}}} & \textbf{SemEv10} & \textbf{SemEv13}  \\
                        & \textbf{AVG}       & \textbf{AVG}        \\
\hline
\cite{amrami-2019}      & 53.6$\pm$1.2     & 37.0$\pm$0.5      \\
\hline
\cite{amrami-2018}      & -         & $25.4 \pm 0.5$      \\
\hline
ELMo                        & 41.8      & 27.6       \\
\hline
BERT                        & 52.0      & 34.5       \\
\hline
XLNet                        & 49.6      & 33.7       \\
\hline
ELMo+embs                        & 45.3      & 28.2       \\
\hline
BERT+embs                        & 53.8      & 36.8       \\
\hline
XLNET+embs                        & 51.1      & 36.1    \\  
\hline
\end{tabular}
\caption{Evaluation on the word sense induction task to the current state-of-the-art models on SemEval2010 and SemEval2013 tasks. 
}
\label{tab:wsi-table}
\end{table}

\section{Extrinsic Evaluation}
\label{sec:extrinsic_eval}

In this section, we show the usefulness of lexical substitution based on neural LMs in the context of two tasks: word sense induction and textual data augmentation. 

\subsection{Word Sense Induction}


WSI is the task of senses identification for a target word given its usages in different contexts. In this problem, we are commonly provided with a corpus of sentences that contain target lemma and part of speech (POS) tag and it's needed to cluster word occurrences, hence, obtaining word senses. For example, suppose that we have the following sentences:

{ \footnotesize
\begin{enumerate2}
    \item He settled down on the river \textbf{bank} and contemplated the beauty of nature.
    \item They unloaded the tackle from the boat to the \textbf{bank}.
    \item Grand River \textbf{bank} now offers a profitable mortgage.
\end{enumerate2}
}
Sentences 1 and 2 must belong to one cluster, but sentence 3 must be assigned to another. This task was proposed in several SemEval competitions \cite{agirre-soroa-2007-semeval,manandhar-etal-2010-semeval,jurgens-klapaftis-2013-semeval}. The current state-of-the-art approach \cite{amrami-2019} relies on substitute vectors, i.e. each word usage is represented as a context-dependent distribution over probable substitutes and clustering is performed over these distributions.

We incorporated an algorithm for word sense induction task in order to compare proposed generators. The algorithm is based on techniques that were described in \cite{amrami-2018,amrami-2019}. On the first step, we generate substitutes for each instance, lemmatize them and take 200 most probable. On the next step we represent these 200 substitutes as a vector by using TF-IDF. Finally, we cluster obtained vectors with agglomerative clusterization with average linkage and cosine distance. 

We evaluate lexical substitutes based on neural LMs in the following datasets: SemEval-2013 and SemEval-2010. We compare our models with the current SOTA on the WSI task -- \cite{amrami-2019}. Table~\ref{tab:wsi-table} demonstrates that combination with embeddings helps to substantially improve generators. For example, a combination of BERT with its embeddings (BERT+embs) improves the results of a BERT model by about 3\% on both data sets. Likewise, a combination of forward LM, backward LM and proximity of ELMo embeddings between substitute and target word, i.e. ELMo+embs generator, raises results on SemEval-2010 task by about 4\%.

\begin{figure}[!h]
    \centering
    \includegraphics[width=0.5\textwidth]{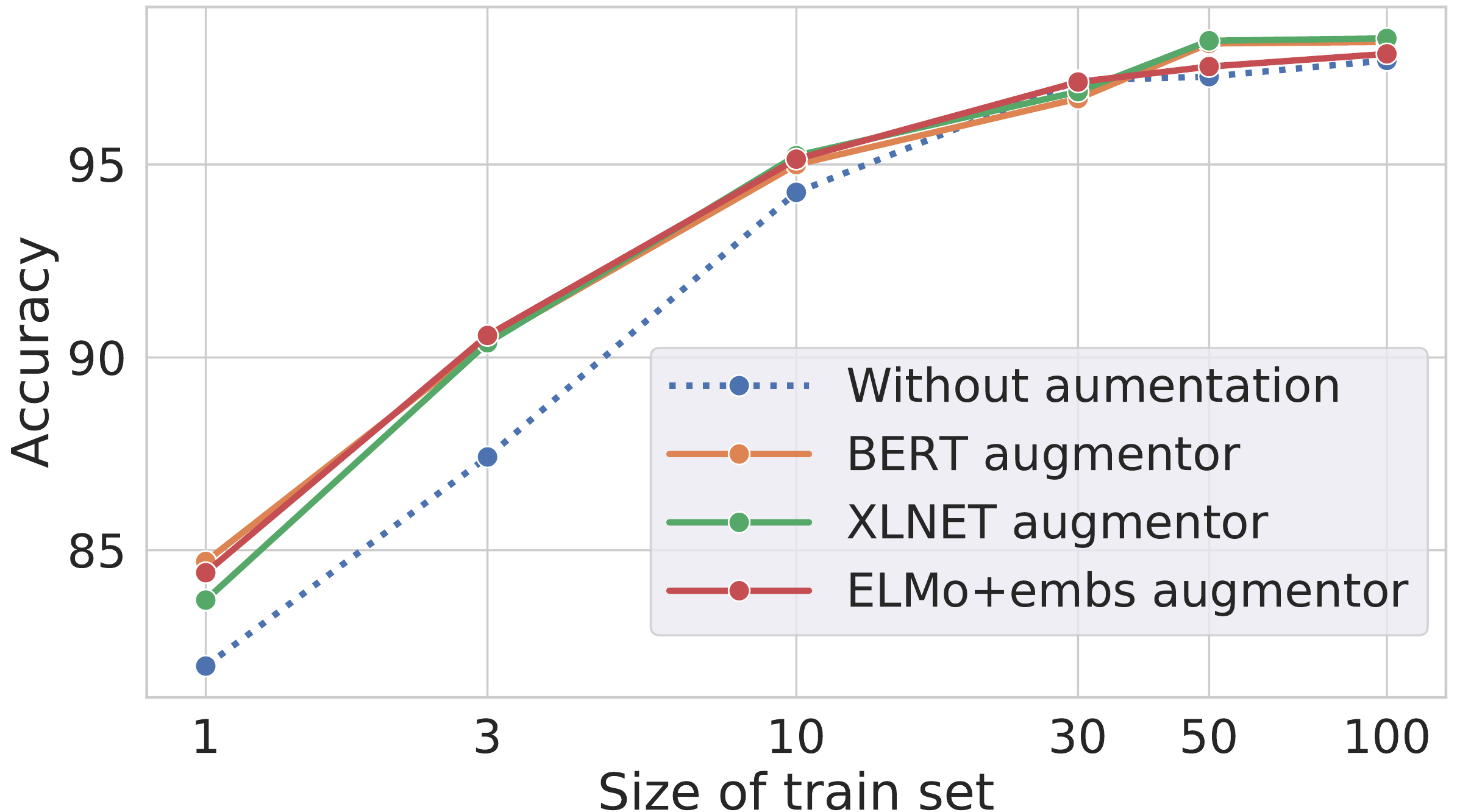}
    \caption{Accuracy on Intent Classification task with different train sizes on SNIPS dataset.}
    \label{fig:aug-accuracy}
\end{figure}


\subsection{Data Augmentation}
Another task that could benefit from contextual substitution is data augmentation. Data augmentation techniques are widely used in computer vision and audio, e.g. image rotation, cropping, etc. 
For textual data, we don't have straightforward techniques for augmentation due to the high complexity of language. There are several papers that address this problem by using contextual substitutions, \cite{kobayashi-2018-contextual,gao-etal-2019-soft,DBLP:journals/corr/abs-1812-06705} to mention a few. Since we can generate substitutes for a word in a sentence, it can be used to create simple paraphrases. 
In this paper, we analyze data augmentation with contextual substitutions on the Intent Classification task.

Intent Classification is necessary for personal digital assistants to decide which action to take in response to some user utterance.
This is essentially a multi-class classification task. When new skills are introduced in assistant, the number of classes grows rapidly. The number of examples for these new classes are usually small, which makes the application of modern deep learning models difficult and requires techniques like data augmentation.

In this paper we use the SNIPS dataset to study how augmentation affects Intent Classification quality. 
The SNIPS dataset \cite{Coucke2018SnipsVP} is a popular public dataset for the Intent Classification and Slot Tagging tasks, which contains 7 intents, 13084/700/700 samples in train/dev/test, respectively. 
Also, SNIPS has a nice feature: it is well balanced by intent.
As a model for the Intent Classification task, we chose the SOTA model on SNIPS --- Capsule NLU which is a capsule-based neural network model \cite{Zhang2018JointSF}. 
We train this model using hyperparameters which were selected in the original paper.

To generate new examples, we use the following algorithm: we select one random word in the sentence corresponding to some slot, next we generate substitutes for this word, and then we sample one substitute with probabilities corresponding to the generated substitutes and replace the original word with the sampled substitute.

\begin{figure}[!h]
\begin{center}

\includegraphics[width=0.5\textwidth]{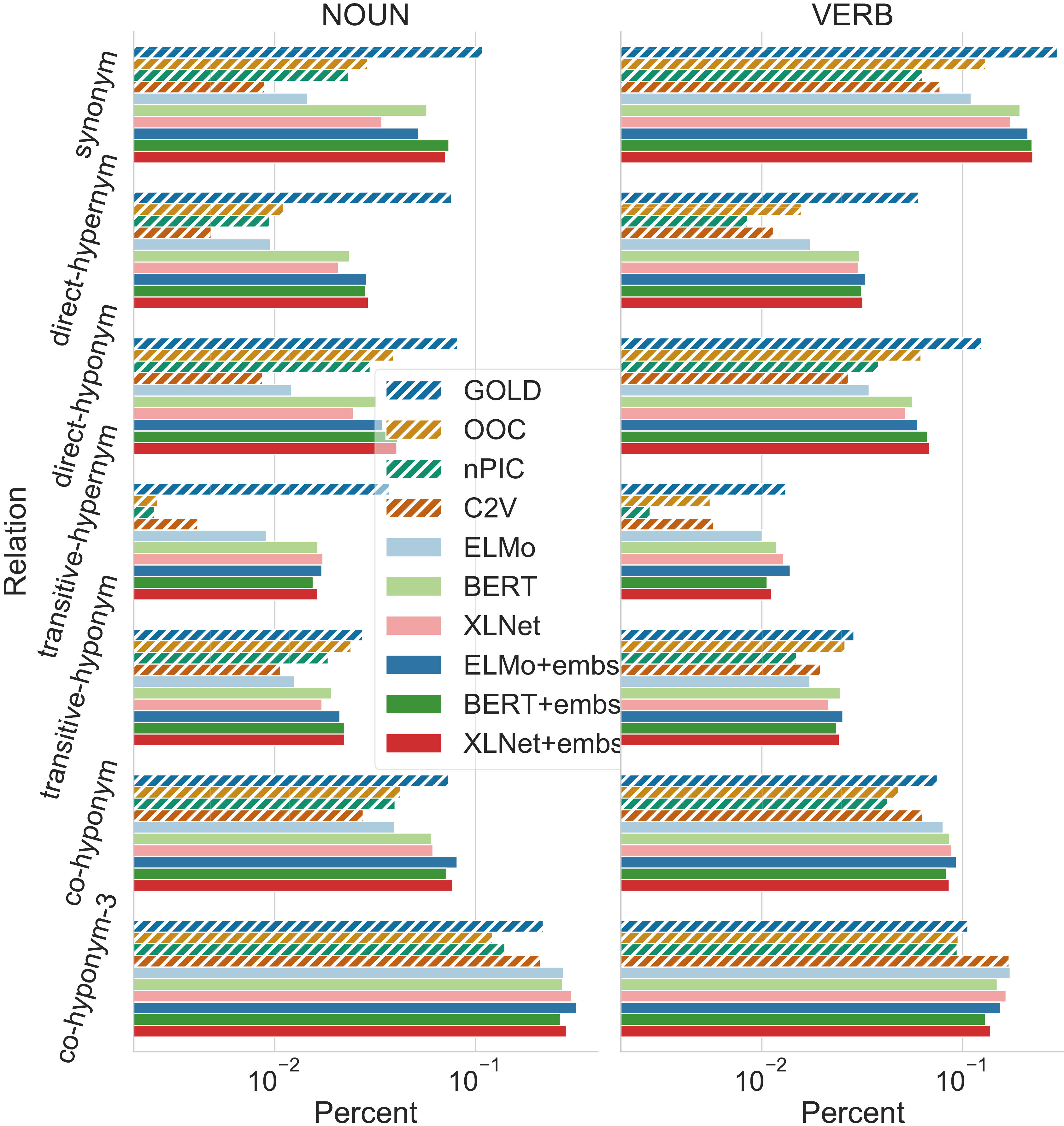}

\caption{Percent of substitutes (log-scale) related to the target by various semantic relations (according to WordNet) on the CoInCo data set.}
\label{fig:relation-types}
\end{center}
\end{figure}

\subsubsection{Results}

We evaluate how data augmentation based on different substitution models affects on Intent Classification 
performance depending on the number of examples in the train set.
For all intents we randomly sampled without replacement the same number of examples ranging from 1\% to 100\%.

In the Figure~\ref{fig:aug-accuracy} we see that the quality of the Intent Classification task begins to sharply decrease when the size of train data reduces to 10\%. Even with 30\% of the train set, it’s enough data to get accuracy score close (0.5\% difference) to the performance on the full data set. Our augmentation allows to improve the quality of Intent Classification. The Figure~\ref{fig:aug-accuracy} shows that augmentation gives a greater increase with a small amount of initial train set (1\%, 3\%, 10\%) than with a larger train set (30\%, 50\%, 100\%).


\begin{figure*}[!h]
\centering
\includegraphics[width=1\textwidth]{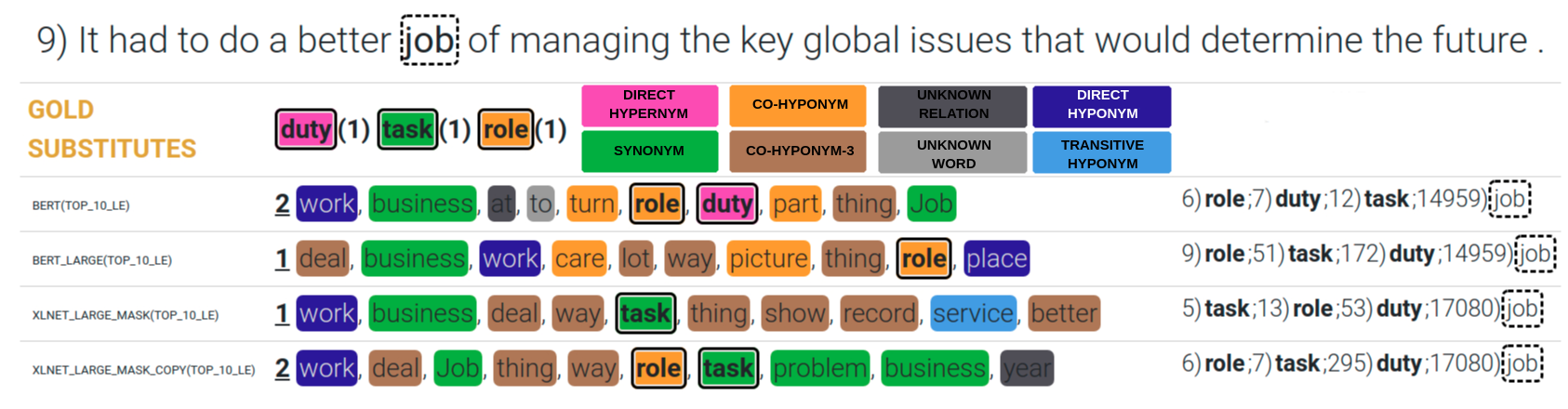}
\caption{Visualisation  interface of various considered neural substitution models facilitating interpretation of the results. The input sentence is placed at the top. The target word is marked by a dashed box. Then gold substitutes follow, their weights are given in brackets. After gold substitutes models predictions come along with the number of true positives to the right of a model name. Each word is colored according to WordNet relation between it and a target word. Here words with a similar to relation highlighted by blue, words with no relation - by red. For each model web application provides ranks of gold substitutes.}
\label{fig:interface}
\end{figure*}

\section{Analysis of Semantic Relation Types}

In this section, we provide an analysis of types of semantic relations produced by various neural language models.

\subsection{Experimental Settings}
We use two lexical substitution corpora this analysis, which were described above: the SemEval 2007 dataset \cite{mccarthy-navigli-2007-semeval} and the CoInCo dataset~\cite{kremer-etal-2014-substitutes}. For each target word and for the substitute we search for two most similar synsets in WordNet~\cite{miller1995wordnet}. Then a relation between these words is identified. If the direct relation is not available we search for a transitive relation: for hypo/hypernyms with no limitation of path length and for co-hyponyms with length of maximum three hops in the graph. We give examples of considered relations in the appendix. Then we count statistics of  relation types. 

For better interpretability of various neural lexical substitution models, we developed a graphical user interface presented in Figure~\ref{fig:interface}. It allows to select the most suitable model based on  interactive processing of user input texts. 

\subsection{Discussion of Results}

Figure~\ref{fig:relation-types} presents results of the experiment. We used several neural language models to show the difference between produced relation types for nouns and verbs. First, as one can observe, for both parts of speech a substantial fraction of words, even produced by the original gold standard annotations has no direct relation to target in terms of WordNet. Also, we note that even human annotators make errors in pos for substitute or a target, e.g. for {\it bright} as an adjective someone gave {\it glitter} as a substitute. For adjectives and adverbs such case takes 15\% and 25\%, and for verbs and nouns less than 7\%. Analyzing substitutes provided by baseline models, OOC and nPIC, we see that unknown word relation prevails taking 40\%. Partly this happens because their vocabularies contain words with typos but we also see that these models don't capture pos of a target word properly for some instances. Proposed models produce much fewer substitutes that are unknown-word according to WordNet for a given pos. BERT and XLNet generate comparable to the gold proportion of such words. This suggests that these models better capture pos tag of a target word and relations between words in a sentence.

Second, for nouns the majority of substitutes fall into either synonyms or (transitive) co-hyponym relation classes. We observe that combinations with embeddings produce consistently more synonyms than corresponding single models, however, still less than humans. When combined with embeddings, BERT and XLNet are on par. Without embeddings BERT outperforms XLNet. If we look at transitive co-hyponyms (co-hyponyms-3 on the figure) we observe the opposite: models combined with embeddings produce fewer substitutes of this type, XLNet outperforms BERT. We hypothesize that the addition of information from embeddings incline models to produce words that are more closely related to a target word as they lie closer to it in a WordNet tree. Analyzing other relations we see the proof to this: the proportion of transitive hypernyms, transitive co-hyponyms and unknown-relation decreases and at the same time proportion of direct hypernyms, direct hyponyms and co-hyponyms increases.

Further, c2v and ELMo without embeddings, which don't see the target, generate the smallest percent of synonyms for all parts of speech except verbs. Also, these models produce much more substitutes with unknown relation to a target word than other models. Combination of these models with embeddings gives rise to all meaningful relations, i.e. co-hyponyms, transitive co-hyponyms, synonyms, etc, as we inject information about a target word.

\section{Conclusion}

In this paper, we presented the first large-scale computational study of three state-of-the-art neural language models (BERT, ELMo, and XLNet) and their variant on the task of lexical substitution in the context. In addition to extensive experimental comparisons on several intrinsic lexical substitution benchmarks, we present a comparison of the models in the context of two applications: word sense induction and text data augmentation. 

Our finding suggests that (i) the simple \textit{unsupervised} approaches based on large pre-trained neural language models yield results comparable to sophisticated traditional \textit{supervised} baseline approaches; (ii) integration of the information about the target substantially boosts the quality of lexical substitution and shall be used whenever possible. 

In addition to comparison on the benchmarks, we also show which models tend to produce semantic relations of which types (synonyms, hypernyms, meronyms, etc.) providing valuable guidelines to practitioners aiming to use lexical substitution in applications. Indeed, depending on the type of semantic relations required in an NLP application one or another type of neural LM shall be used.

\bibliography{acl2020}
\bibliographystyle{acl_natbib}

\end{document}